\documentclass[letterpaper, 10 pt, conference]{ieeeconf}

\IEEEoverridecommandlockouts                              %

\usepackage{times}

\usepackage{multicol}
\usepackage{booktabs}
\usepackage[bookmarks=true]{hyperref}
\usepackage{amsmath}
\usepackage{amsfonts}
\usepackage{physics}
\usepackage{graphicx}
\usepackage{color}
\usepackage{subcaption}
\usepackage{comment}
\usepackage[ruled,noend]{algorithm2e}
\usepackage{algpseudocode}
\usepackage{float}
\usepackage{tikz}
\usetikzlibrary{positioning}
\usetikzlibrary{calc}

\usepackage[all]{nowidow}

\graphicspath{{figures/}}

\newcommand{\m}{\mathbf{m}}
\newcommand{\p}{\mathbf{p}}

\newcommand{\x}{\mathbf{x}}

\newcommand{\0}{\mathbf{0}}
\newcommand{\y}{\mathbf{y}}

\newcommand{\cc}{c}
\newcommand{\cz}{\cc_\emptyset}
\newcommand{\mankin}{\mathbf{f}}
\newcommand{\objkin}{\mathbf{g}}
\newcommand{\segroup}{\mathrm{SE}(3)}
\newcommand{\trajectoryobjective}{\mathcal{J}}

\makeatletter
\newcommand{\showfont}{encoding: \f@encoding{},
  family: \f@family{},
  series: \f@series{},
  shape: \f@shape{},
  size: \f@size{}
}

\begin{document}

\author{Jimmy Envall$^{1}$, Roi Poranne$^{2}$, Stelian Coros$^{1}$
\thanks{$^{1}$ The authors are with the Department of Computer Science, ETH, Zurich, Switzerland.
{\tt\small jimmy.envall@inf.ethz.ch; scoros@inf.ethz.ch}}%
\thanks{$^{2} $Roi Poranne is with the Department of Computer Science, University of Haifa, Haifa, Israel.
{\tt\small roiporanne@cs.haifa.ac.il}}%
\thanks{This work was supported by a ETH RobotX research grant funded
through the ETH Zurich Foundation.}
}

\title{Differentiable Task Assignment and Motion Planning}

\maketitle

\begin{abstract}
Task and motion planning is one of the key problems in robotics today.
It is often formulated as a discrete task allocation problem combined with continuous motion planning. 
Many existing approaches to TAMP involve explicit descriptions of task primitives that cause discrete changes in the kinematic relationship between the actor and the objects.
In this work we propose an alternative, fully differentiable approach which supports a large number of TAMP problem instances.
Rather than explicitly enumerating task primitives, actions are instead represented implicitly as part of the solution to a nonlinear optimization problem.
We focus on decision making for robotic manipulators, specifically for pick and place tasks, and explore the efficacy of the model through a number of simulated experiments including multiple robots, objects and interactions with the environment.
We also show several possible extensions.
\end{abstract}

\IEEEpeerreviewmaketitle

\section{Introduction}

To perform manipulation tasks, even simple ones such as picking up an object, robots must solve two tightly coupled problems.
The first is decision making: what strategy should be used to pick the object?
The second is motion planning: how to move and perform the picking motion in an optimal way based on the selected strategy?
These problems together are a simplistic instance of the \emph{task and motion planning} problem, or TAMP.
Another instance of the problem involves several robotic manipulators that are tasked with sorting and organizing a set of objects that are scattered around their environment.
One way to solve this problem is to assign each robot to specific pick-and-place tasks, and then find the optimal trajectories for all of them simultaneously.
The main question TAMP aims to address is, what would be the optimal assignment?
The challenge there stems from the interplay between the two problems that make up TAMP: task planning, and motion planning.

The cost of a motion plan given a specific task is hard to predict and expensive to evaluate, and even the smallest change to the task description can cause the motion plan to become infeasible.
Thus, TAMP approaches are often concerned with finding efficient ways for searching in the space of task assignments \cite{bib:lozano,Toussaint2015,Toussaint2017}.

The combinatorial complexity of TAMP could be partly mitigated if task planning could be formulated in a unified, continuous way.
Furthermore, and perhaps more importantly, a continuous formulation would allow the problem to be integrated in differentiable simulators and neural networks and ultimately be a step toward high-level decision making in complex situations.
For these reasons, our goal in this paper is to propose a fully continuous formulation applicable to a considerable subset of the TAMP problem domain including task assignment and motion planning.
Our emphasis is on decision-making for robotic manipulators, particularly for pick-and-place tasks.
We address this challenge by treating task assignment in an \emph{implicit} manner.
The idea is, instead of assigning a pick-and-place task, to associate robots with objects using time-dependent, real functions.
These functions, in some sense, express for each point in time a degree of which a specific robot should hold a specific object.
Hence, in contrast to explicitly defined pick events and place events, they only \emph{implicitly} define them.

Based on this concept, we define a smooth optimization problem that can be readily solved using gradient-based methods, such as Newton's method.
We demonstrate the potential of this approach in a variety of settings, including multiple manipulators and objects.
We also show that this approach is easily extendable to handle different grasps, handovers and more complex interactions.
\begin{figure}
    \centering
    \includegraphics[width=\columnwidth]{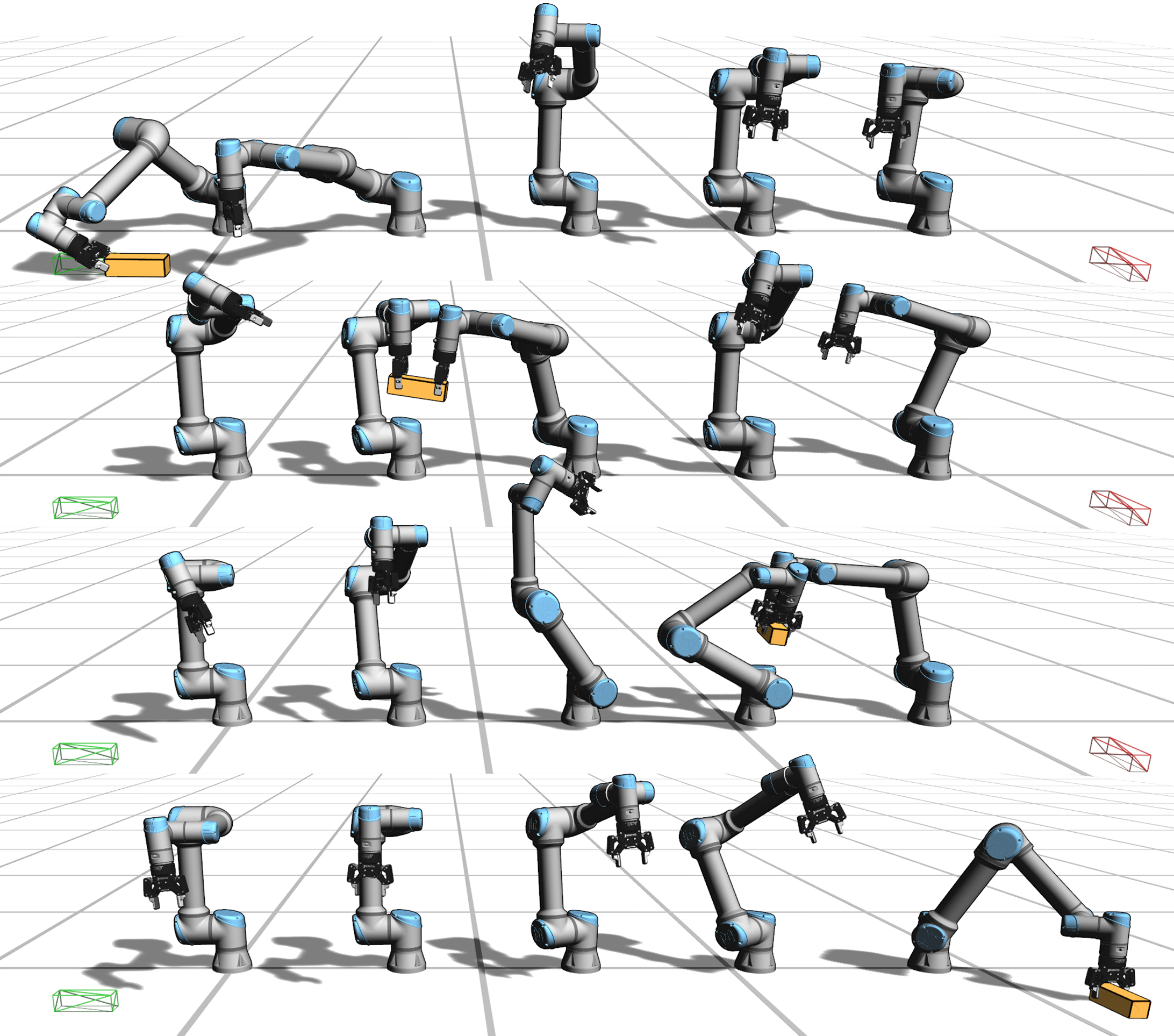}
    \caption{The number of available actions, such as handovers, increases exponentially with the number of manipulators.}
    \label{fig:teaser}
    \vspace{-1.5em}
\end{figure}
\section{Related work}
Modern approaches for task and motion planning combine discrete high level decision making with continuous parameter search in different ways. In \cite{bib:lozano} the configuration space is discretized and combined with the action skeleton into a discrete constraint satisfaction problem that is then handed to a generic CSP solver. Another popular method known as Logic Geometric Programming (LGP) \cite{Toussaint2015,Toussaint2017} combines symbolic action search with %
a non-linear program for finding the trajectory.
The exponential growth of the discrete action space poses challenges for longer planning horizons, for which several extensions have been developed \cite{Driess2019,Braun2021}.

Other approaches, e.g. \cite{bib:thomason_unified_2019,Garrett2018} suggest to find the trajectory by sampling.
In \cite{bib:thomason_unified_2019} the symbolic actions and the continuous parameter spaces are fused and solved simultaneously using an off-the-shelf motion planner while in \cite{Garrett2018} the domain specific constraints are utilized for factoring the problem, enabling efficient sampling.
As can be inferred by \cite{Garrett2021}, %
the main differences between different approaches are how they solve for the continuous variables and how constraint satisfaction interacts with the search for high level action sequences.
In general, TAMP is a very active research topic, and the review in \cite{Garrett2021} is highly recommended.

TAMP covers a wide range of problems \cite{bib:lagriffoul2018platform}, and successful completion of a task may hinge on effective manipulation, which is an active research field on its own \cite{bib:billard2019trends,bib:kleeberger2020survey,bib:SUOMALAINEN2022104224}.
However, while task planning for a single manipulator already is challenging, increasing the number of agents puts additional emphasis on the planning and decision making algorithms.
In some settings, e.g. in a factory or in a warehouse, less general algorithms may be sufficient.
Pick and place problems involving multiple robots and handovers may be solved using heuristics and sampling for exploring the search space \cite{bib:8901083}.
Another option might be to consider the individual robots in a non-cooperative fashion \cite{bib:BOZMA2012530}.

Planning through kinematic modes using optimization has previously been studied in \cite{bib:mordatch_discovery_nodate} and \cite{bib:mordatch_contact-invariant_2012} in the context of motion synthesis for computer graphics and in \cite{bib:posa2014direct} for trajectory planning of rigid body systems. Similarly to our work, these contributions also rely on optimization and additional variables for describing switches between different modes. Sampling based methods include e.g. \cite{bib:hauser_2010} which extends the popular probabilistic roadmaps method \cite{bib:kavraki1996probabilistic} with multi-modal capabilities.

A similar concurrent work found in \cite{bib:takano_continuous_2021} builds upon a concept known as signal temporal logic (STL) and a smooth approximation thereof \cite{bib:pant_smooth_2017}, which enables a fully continuous approximation to task planning.
The formulation proposed in \cite{bib:takano_continuous_2021} also features auxiliary variables similar to the ones used in this work. %
However, the formulation relies on constraints that scale quadratically with the number of objects.
Additionally, it was not demonstrated on multiple robots, and does not include a grasping parametrization and uses floating end effectors during the planning with joint angles computed in a separate step.

The idea of treating naturally discrete phenomena as continuous has also been employed in other fields, such as topology optimization.
A popular method for optimizing structural integrity is known as Solid Isotropic Micro-Structure with Penalization, or SIMP \cite{bib:bendsoe1989optimal,bib:rozvany2001aims}.
In SIMP the target domain is divided into finite elements whose occupancy is modeled using a continuous range of values.
For non-porous materials fractional values are difficult to interpret, and therefore penalization techniques are needed to drive the occupancy to either empty or solid.
However, in our method, fractional values do not constitute an issue.

\section{Overview}\label{sec:overview}
\begin{figure}
    \centering
    \includegraphics[trim={0 0 0 0}, clip, width=\columnwidth]{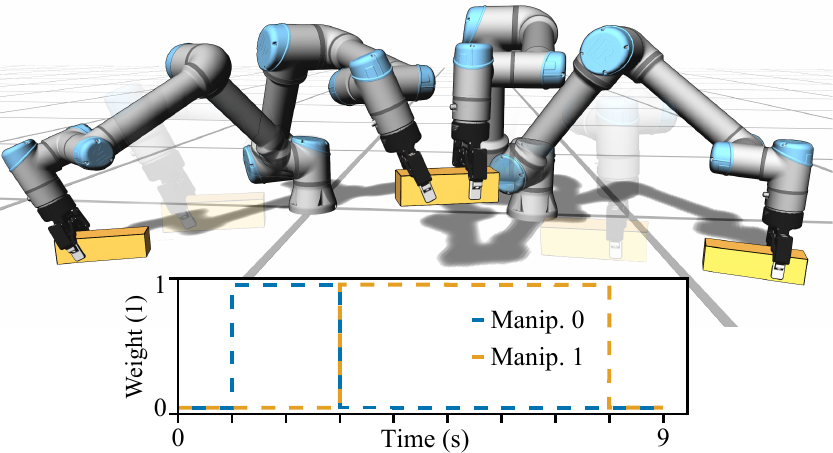}
    \caption{Example setup with two robotic manipulators and one movable object.}
    \label{fig:overview_setup}
    \vspace{-1.5em}
\end{figure}
The archetypal trajectory optimization problem we build upon is
\begin{equation}\label{eq:trajectory_optimization}
\begin{aligned}
\min_{\x(t)} \quad & \trajectoryobjective(\x(t))\\
\textrm{s.t.} \quad & \cc(\x(t)) = 0,
\end{aligned}
\end{equation}
where $\x(t)$ denotes the trajectory of all state variables, $\trajectoryobjective$ is the objective, e.g. shortest path, and $c(\x)$ denotes user-specified constraints such as initial and target positions on the trajectory.
To begin our discussion, consider a scenario where a robotic manipulator is holding a box.
The final desired placement of the box is out of reach for the robot that is currently holding it, but there is another robot close by that can take the box from the first robot and bring it to its goal.
This handover maneuver is a decisive high-level task that needs to be actively chosen.
An illustration is shown in \ref{fig:overview_setup}.

The pose of the object can be seen as a link in the kinematic chain.
The pose is thus governed by the kinematic equations that are induced by the robot holding it.
Put differently, there is a set of constraints that the box and the robot holding it must satisfy.
During a handover, this set of constraints change in what is known as a \emph{kinematic switch}.

Much of the previous work mentioned above deals with the explicit sequencing of actions.
We instead enable the kinematic switches to emerge as a result of solving a nonlinear program.
This is made possible by mollifying the switches.
This way we can simultaneously find the action sequence and the motion trajectory using only gradient-based methods without relying on integer-based techniques.

This example can be expressed as an extension of \eqref{eq:trajectory_optimization}:
\begin{equation*}\label{eq:simplified_problem}
\begin{aligned}
\min_{\x(t)} \quad & \trajectoryobjective(\x(t))\\
\textrm{s.t.} \quad & \cc(\x(t)) = 0\\
  &\cc_1(\x(t)) = 0 \lor \cc_2(\x(t)) = 0 \lor \cz(\x(t)) = 0\; \forall\; t
\end{aligned}
\end{equation*}
where $\x(t)$ also includes the trajectory of the object.
The constraints $\cc_1(\x)$ and $\cc_2(\x)$ define the grasping relationship between the object and each manipulator, and $\cz$ is a function whose value is zero when the object rests on the ground.
The distinguishing feature of this problem is that only one of these constraints needs to be satisfied at a given time.
In other words, the box must be held by one or both manipulators and/or be on the ground, but it cannot float.

While this could potentially be formulated as a mixed integer problem, we instead opt for a fully continuous formulation which we detail in the next section.
Our main idea is to \emph{relax} the kinematic switches in the problem by introducing time-dependent \emph{association} weights $w_i(t):[0,T]\to [0,1],\; i\in\{1,2\}$.
These weights signify the association between each robot and the box, or "how much" the constraints $c_1$ and $c_2$ need to be satisfied.
With these weights, the optimization problem is transformed into a complementarity problem:
\begin{equation}\label{eq:simplified_problem_w_constraints}
\begin{aligned}
&\min_{\x, w_1, w_2} \quad &&\trajectoryobjective(\x(t))\\
&\,\,\,\,\,\textrm{s.t.} \quad  && c(\x(t)) = 0 \\
&\,                   && w_i(t) c_i(\x(t)) = 0, \forall\; i\in\{1,2\}\\
&\,                   && w_\emptyset(t) \cz(\x(t))  = 0                   \\
\end{aligned}
\end{equation}
where $w_\emptyset(t)=1-\sum_i w_i(t)$.
Any of the constraints $c_1,c_2$ and $\cz$ can now become inactive by choosing the weights $w_1$ and $w_2$ appropriately.

\section{Method}\label{sec:detailed_method}
The problem presented in \eqref{eq:simplified_problem_w_constraints} provides the foundation of our approach.
In this section we refine the formulation to add support for task allocation, multiple objects and robotic manipulators, and more.

\noindent\textbf{Trajectories}
We denote the joint angles of a manipulator $i\in\mathcal{M}$ at time $t\in [0,T]$ by $\m_i(t)\in \mathbb{R}^{n}$.
The joint angles can be subjected to box constraints of the form $\m^{\text{min}}_i \leq \m_i(t) \leq \m^{\text{max}}_i$ that correspond to the physical joint limits of the robot.
The trajectory of an object $j\in\mathcal{P}$ is written analogously as $\p_j(t)\in\mathbb{R}^6$.
Objects consist of only one rigid body and $\p_j$ thus directly describes an object's pose in world coordinates.
We use $\mankin: \mathbb{R}^{n}\to \segroup$ for denoting the forward kinematic function that maps the manipulator state $\m_i$ to an end effector position and orientation in the world coordinate frame.

\noindent\textbf{Task selection}
The allocation of a task $j$ to a manipulator $i$ is modeled using a function $w_{ij}(t):[0,T]\to [0,1]$ as outlined in \ref{sec:overview}.
These variables form a key component of the path constraints that will be developed in \ref{sec:kinematic_switches}.

\noindent\textbf{Resting}
Objects may need to be released before they have reached the goal pose.
To ensure physically plausible behavior we restrict the resting pose of the object to be in a set of stable configurations.
E.g. a cube can be placed with any face towards the ground, implying that the elevation of the cube as well as one rotational axis is fixed.
We introduce the functions $r_{dj}(t): [0,T]\to [0,1]$ where $d$ corresponds to a specific resting pose.
We use these functions to construct the resting orientation constraints in \ref{sec:resting_constraints}.

\noindent\textbf{Grasping}
During handovers from one robot to another, different grasping poses may be necessary in order to prevent robots from colliding.
In this work we use cuboid objects of size $0.06\text{ m}\cross 0.06\text{ m}\cross 0.2\text{ m}$.
We model the grasp pose with two degrees of freedom; one along the longitudinal axis of the object and one being the angle between the gripper and the local y-axis of the object.
A similar parametrization has been used in \cite{bib:zimmermann_trilevel} with an additional degree of freedom.

We denote the longitudinal offset and the angle with $\delta_{ij}$ and $\theta_{ij}$ respectively.
The longitudinal offset is subject to a box constraint that depends on the physical dimensions of the object, i.e. $\delta^{\text{min}}_{j}\leq \delta_{ij} \leq \delta^{\text{max}}_{j}$.
We note that the grasping parameters are defined per manipulator-object pair and therefore are independent of $t$.

For conciseness we hereafter write $\y_{ij}(t)=\begin{bmatrix}w_{ij}(t) & \delta_{ij} & \theta_{ij}\end{bmatrix}^T$.
Given the grasping parameters and the object state $\p_j$ we can define $\objkin: (\p_j,\y_{ij})\to \segroup$ that maps the object state and grasping parameters to a grasping pose for the end effector in world coordinates.

\subsection{Differentiable kinematic switches}\label{sec:kinematic_switches}
We model the kinematic switches as path constraints that depend on $\m$, $\p$ as well as the auxiliary functions $w_{ij}(t)$.
The constraint has the form
\begin{equation}\label{eq:position_constraint}
    C_\text{pos}:= w_{ij}
\left(
\mankin_i(\m_i) - \objkin(\p_j,\y_{ij})
\right)
= \0
\end{equation}
for manipulator $i$ and object $j$.
In addition we constrain the velocities of the non-actuated bodies such that they equal the weighted sum of the velocities of the end effectors that are currently moving it. The velocity constraint is of the form
\begin{equation}\label{eq:velocity_constraint}
    C_\text{vel}:=
\dot{\objkin}(\p_j,\y_{ij}) - \sum_{i\in\mathcal{M}} w_{ij}\dot{\mankin}_i(\m_i)
= \0.
\end{equation}
This constraint also ensures that the velocity of an object is zero when $w_{ij}=0\;\forall\;i\in\mathcal{M}$.
Together with \eqref{eq:position_constraint}, \eqref{eq:velocity_constraint} ensures that an object will be moving only when a manipulator holds it.
These two constraints thus fully describe the dynamical relationship between the manipulator and the objects to be manipulated.

\subsection{Resting constraints}\label{sec:resting_constraints}
For a cube or a block with six faces the resting constraint can be expressed as a constraint on the elevation and on one of the orientation axes of the object.
The number of constraints thus equals the number of available resting orientations.
For brevity we write $\hat{w}_{dj}(t)=1-\sum_{i\in{\mathcal{M}}}w_{ij}(t)$.
The resting constraint can then be expressed as
\begin{equation*}
    C_\text{rest}:= \hat{w}(t)_{dj} r_{dj}(t)\phi_d(\objkin(\p_j,\y_{ij})) = \0
\end{equation*}
where $\phi_d$ is a function that captures the difference between the relevant rotation axis for the resting pose $d$ as well as the resting elevation, e.g. corresponding to a floor or a table. By setting $\sum_dr_{dj}(t)=1$ we ensure that at least one constraint is active when $\hat{w}(t)_{dj}>0$.

\subsection{Collisions}\label{sec:collision_constraints}
Collision free trajectories are ensured by imposing a collision constraint of the form
\begin{equation*}
    C_\text{collision} := c_\text{collision}(\mathcal{K}_a(\m,\p),\mathcal{K}_b(\m,\p)) \geq 0\quad\forall\quad a,b\in\mathcal{C}
\end{equation*}
where $\mathcal{K}_a,\mathcal{K}_b$ denote a pair of forward kinematic functions for collision primitives $a,b\in\mathcal{C}$ in the scene and the value of $c_\text{collision}$ is proportional to the squared distance between the collision primitives and $\m,\p$ denote the stacked trajectory vectors. We refer the reader to \cite{bib:dca} for details.

\subsection{Trajectory optimization problem}
By stacking the manipulator and object trajectories as well as the grasping parameters into $\m$, $\p$ and $\y$ respectively we can now write the trajectory optimization problem as
\begin{equation}\label{eq:trajectory_problem}
\begin{aligned}
\min_{\m,\p,\y} &&\trajectoryobjective(\m,\p,\y) \\
\textrm{s.t.}  && C_\text{pos} &= 0\;&&\forall\; i\in\mathcal{M},\; j\in\mathcal{P} \\
  &&C_\text{vel} &= 0\;&&\forall\; j\in\mathcal{P} \\
  &&C_\text{rest} &= 0\;&&\forall\; j\in\mathcal{P}, \\
  &&C_\text{collision} &\geq 0\;&&\forall\; a,b\in\mathcal{C} \\
  &&\m^{\text{min}}_i \leq \m_i &\leq \m^{\text{max}}_i &&\forall\; i\in\mathcal{M} \\
  &&\delta^{\text{min}}_j \leq \delta_{ij} &\leq \delta^{\text{max}}_j &&\forall\; i\in\mathcal{M},\; j\in\mathcal{O} \\
\end{aligned}
\end{equation}
where $\trajectoryobjective$ is an objective function.
In the experiments we also penalize joint velocities and accelerations of the manipulators, i.e.
\begin{equation*}
    f_v(\m_i):= \|\dot{\m}_i\|^2
\end{equation*}
and
\begin{equation*}
    f_a(\m_i):=\|\ddot{\m}_i\|^2
\end{equation*}
as well as the end effector velocities
\begin{equation*}
    f_{\text{ee}}(\m_i):=\|\dot{\mankin}_i(\m_i)\|^2.
\end{equation*}
$\trajectoryobjective$ then reads
\begin{equation*}
    \trajectoryobjective = \sum_{i\in\mathcal{M}} \left(\beta_1f_v(\m_i) + \beta_2f_a(\m_i) + \beta_3f_{\text{ee}}(\m_i) \right)
\end{equation*}
where $\beta_l\in\mathbb{R},l\in\{1,2,3\}$ are constant weights.

The constraints in \eqref{eq:trajectory_problem} are sufficient for preventing a single manipulator from holding multiple objects simultaneously. However, in our numerical experiments we have found that adding an additional constraint that limits the capacity of the manipulators is beneficial for guiding the optimization.
To prevent manipulators from acting on multiple objects simultaneously we introduce a capacity constraint of the form
\begin{equation*}\label{eq:capacity_constraint}
    C_\text{cap}:=\sum_{j\in\mathcal{P}}w_{ij}(t) \leq 1\;\forall \; t\in [0,T],
\end{equation*}
which prevents a single manipulator from being fully responsible for multiple objects simultaneously.

\subsection{Constraint manifold}
\label{sec:constraint_manifold}
At the core of our formulation are the position and velocity constraints \eqref{eq:position_constraint}\eqref{eq:velocity_constraint}.
Their most important property becomes apparent only when the constraints are considered over time.

In order to illuminate this property we study the constraints in a one dimensional setting with one free manipulator located at $m: t\rightarrow \mathbb{R}$ and a point object located at $p: t\rightarrow \mathbb{R}$ and a constant weight $w$.
The position and velocity constraints then reduce to
\begin{align*}
    w\left(m(t)-p(t)\right) = 0
\end{align*}
and
\begin{align}
\label{eq:1d_velocity_constraint}
    \dot{p}(t)-w\dot{m}(t) = 0
\end{align}
respectively. The constraint violation over time $t\in [t_1,t_2]$ can be quantified using the $L^2$ norm:
\begin{align}
\label{eq:1d_manifold}
    d=\int_{t_1}^{t_2}\left(w\left(m(t)-p(t)\right)\right)^2+\left(\dot{p}(t)-w\dot{m}(t)\right)^2dt.
\end{align}

\begin{figure}
    \centering
    \includegraphics[width=\columnwidth]{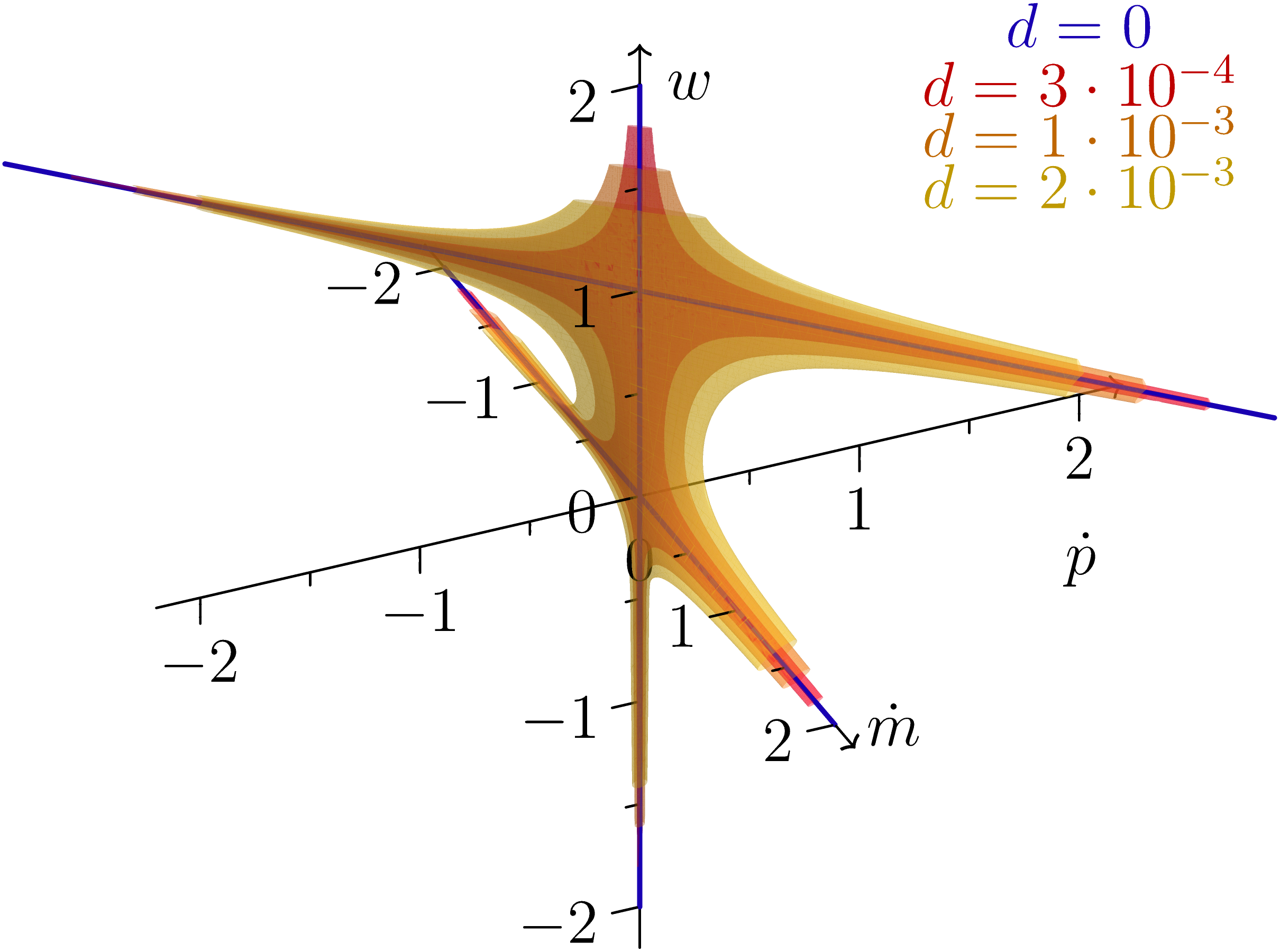}
    \caption{Level sets of \eqref{eq:1d_manifold}. Non-zero velocities of a object-manipulator pair are feasible simultaneously only for $w=1$.}
    \label{fig:1d_constraint_levels}
    \vspace{-1em}
\end{figure}

From \eqref{eq:1d_manifold} we can deduce that:
\begin{enumerate}
    \item If $w=0$, the terms corresponding to the position constraint and the manipulator velocity vanish, and thus it must hold that $\dot{p}(t)=0$.
    \item If $w\neq 0$, the position of the manipulator and the object must coincide over the whole time span for the first term to vanish. Since the position of the manipulator and the object must coincide over time, the velocity of the manipulator and the object must be the same. Thus, in order for the velocity terms to cancel out, it must either hold that $w=1$ or $\dot{p}(t)=\dot{m}(t)=0$.
\end{enumerate}
This property can be visualized using the level sets of \eqref{eq:1d_manifold} when $m(t_1)=p(t_1)$ (Fig.  \ref{fig:1d_constraint_levels}).

Every additional manipulator will add a position constraint as well as a velocity term of the form $-w_i\dot{m}_i(t)$ to \eqref{eq:1d_velocity_constraint}. Thus, when multiple manipulators are moving the object, all of their velocities must be equal to the velocity of the object, and for the (non-zero) velocity terms to cancel out it must hold that $\sum_iw_i=1$. Due to the shape of the constraint manifold, the sum of the weights associated with one object will tend towards either 0 or 1. Individual weights between 0 and 1 appear when multiple robots are transporting the same object simultaneously. Our formulation should therefore not be interpreted as a relaxation of the corresponding mixed integer problem where the weights are binary.

\subsection{Nonlinear program}
We transform \eqref{eq:trajectory_problem} into a nonlinear program in order to solve it numerically.
We model the trajectories using cubic Hermite splines on the $N-1$ segments that are defined by $N$ time points.
For $w_{ij}$ and $r_{dj}$ we use piece wise constant functions defined on $N-1$ segments.
Finally we convert the constrained nonlinear program to an unconstrained problem by applying quadratic and cubic penalty functions to the equality and inequality constraints respectively and solve it using the Gauss-Newton method.
We use CHOLMOD \cite{bib:cholmod} for solving the emerging linear system and \cite{bib:dca} for collision avoidance.
The position and velocity constraints \eqref{eq:position_constraint}\eqref{eq:velocity_constraint} are evaluated at the end points as well as the midpoint of every segment, and the collision constraints at 11 equally spaced points on each segment in order to provide sufficient coverage.

The benefit of the continuous association weights detailed in \ref{sec:detailed_method} can be demonstrated experimentally.
Consider a setup containing two objects that need to be picked up and placed in a different position on the floor, and one manipulator capable of moving the objects.
Solving the optimization problem results in a schedule for moving the blocks.
\begin{figure}
    \centering
    \begin{tikzpicture} 
        \draw (0, 0) node {\includegraphics[width=3.25in]{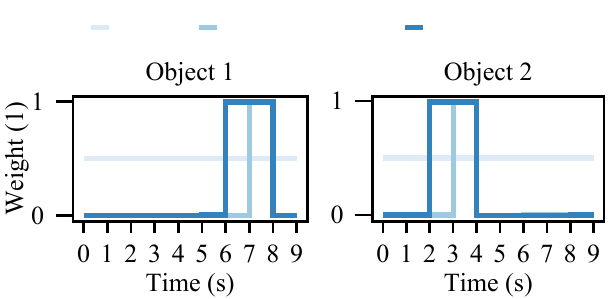}};
        \draw (0.68, 1.7) node[align=center] {Initial\hspace{0.65cm}Solution with $\mathcal{J}$\hspace{0.4cm}Solution with $\hat{\mathcal{J}}$};
    \end{tikzpicture}
    \caption{The association weights for a manipulator moving two objects. The solutions are obtained before and after adding $f_a(\p_1)$ and $f_a(\p_2)$ to the objective function.}
    \label{fig:ablation_weights_control}
    \vspace{-1em}
\end{figure}

Next we add two additional terms to the objective that captures the acceleration of the objects.
We continue the optimization with the updated objective ${\hat{\mathcal{J}}:=\mathcal{J} + f_a(\p_1)+f_a(\p_2)}$.
The resulting schedule is shown in Fig. \ref{fig:ablation_weights_control} with a trajectory length of 9 segments.
As can be seen, the new terms influence the length of the time windows during which the objects are moving.
This shows that the emerging schedule is directly affected by the objective function.

\section{Results}\label{sec:results}
We evaluate the formulation \eqref{eq:trajectory_problem} by applying it to a number of scenarios in a simulated environment.
The experiments were implemented in C++ and executed on a desktop computer equipped with an 16-core AMD Ryzen 5950X 3.4 GHz CPU and 32 GB of RAM.
All tasks are designed such that they can be solved by the robots in the scene.
The association weights $w_{ij}$ are initialized to 0.5 unless stated otherwise.

\subsection{Environmental influence}
This experiment demonstrates how the allocation of objects to manipulators is affected by obstacles in the environment.
The scene consists of one fixed UR5 and a Kinova mounted on an omnidirectional wheeled platform.
We use a trajectory consisting of 9 segments, i.e. 10 discrete time points where the joint values of the manipulators and root position of the Kinova platform are initialized to a resting state.
The trajectory of the block is initialized such that the final state corresponds to the goal pose while all other states are initialized to the starting pose.

\begin{figure}
    \begin{subfigure}[b]{0.33\columnwidth}
    \centering
        \begin{tikzpicture}[every node/.style={inner sep=0,outer sep=0}]
            \draw (0, 0) node {\includegraphics[width=2.8cm]{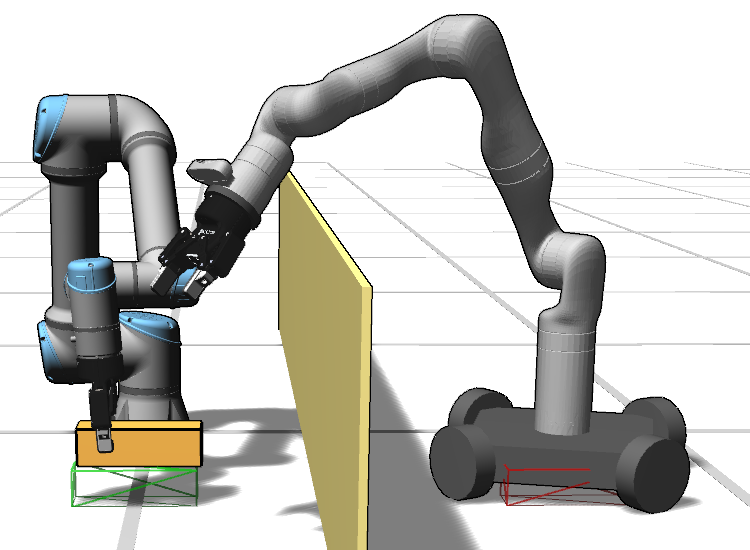}};
            \draw (0, 1.15) node[align=center] {1.4 s};
        \end{tikzpicture}
    \end{subfigure}%
    \hfill
    \begin{subfigure}[b]{0.33\columnwidth}
    \centering
        \begin{tikzpicture}[every node/.style={inner sep=0,outer sep=0}]
            \draw (0, 0) node {\includegraphics[width=2.8cm]{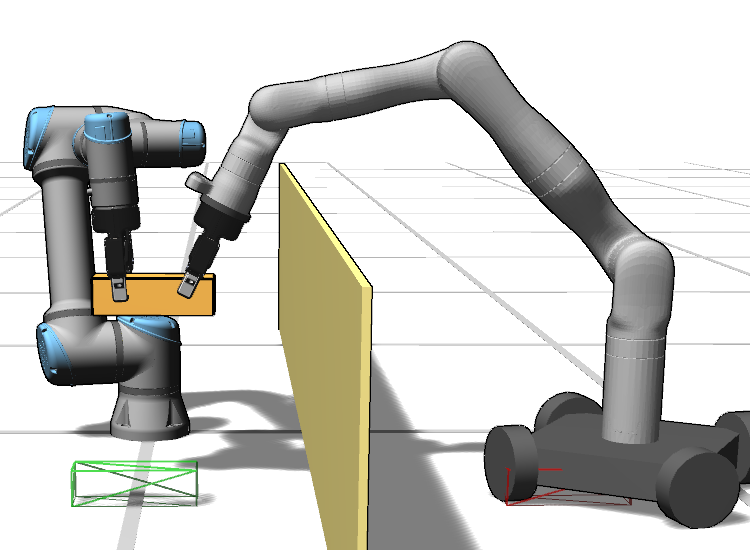}};
            \draw (0, 1.15) node[align=center] {2 s};
        \end{tikzpicture}
    \end{subfigure}%
    \hfill
    \begin{subfigure}[b]{0.33\columnwidth}
    \centering
        \begin{tikzpicture}[every node/.style={inner sep=0,outer sep=0}]
            \draw (0, 0) node {\includegraphics[width=2.8cm]{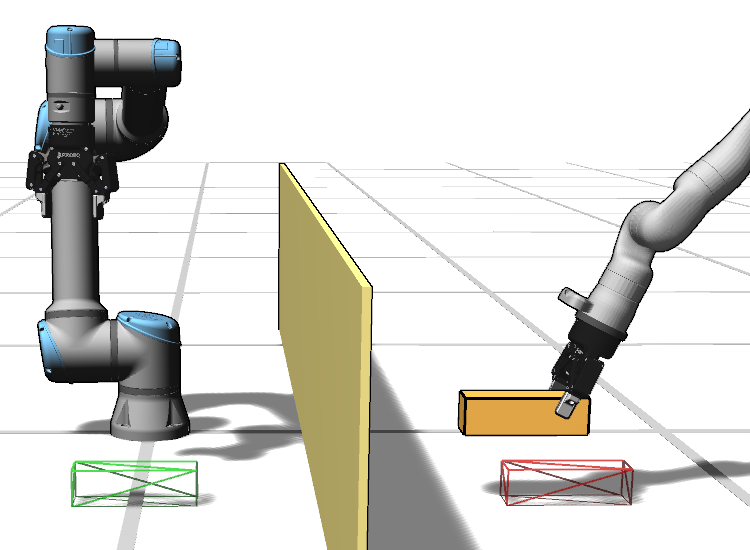}};
            \draw (0, 1.15) node[align=center] {7 s};
        \end{tikzpicture}
    \end{subfigure}
    \par\bigskip
    \begin{subfigure}[b]{0.33\columnwidth}
    \centering
        \begin{tikzpicture}[every node/.style={inner sep=0,outer sep=0}]
            \draw (0, 0) node {\includegraphics[width=2.8cm]{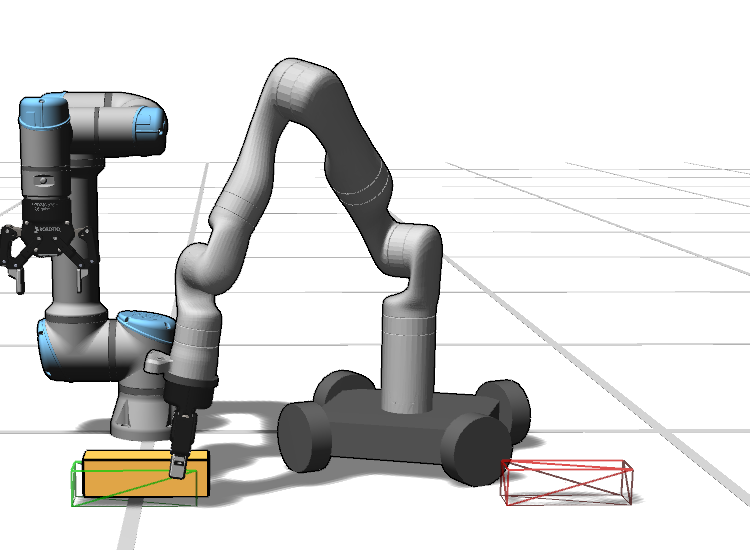}};
            \draw (0, 1.0) node[align=center] {1.4 s};
        \end{tikzpicture}
    \end{subfigure}%
    \hfill
    \begin{subfigure}[b]{0.33\columnwidth}
    \centering
        \begin{tikzpicture}[every node/.style={inner sep=0,outer sep=0}]
            \draw (0, 0) node {\includegraphics[width=2.8cm]{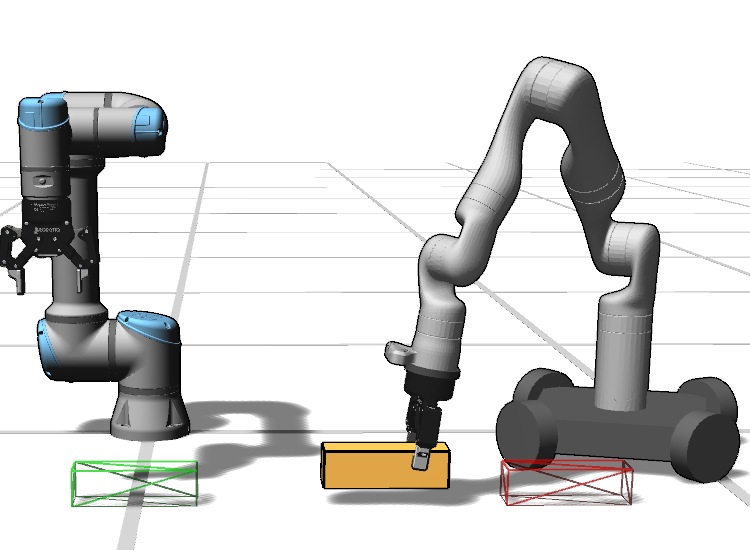}};
            \draw (0, 1.0) node[align=center] {5 s};
        \end{tikzpicture}
    \end{subfigure}%
    \hfill
    \begin{subfigure}[b]{0.33\columnwidth}
    \centering
        \begin{tikzpicture}[every node/.style={inner sep=0,outer sep=0}]
            \draw (0, 0) node {\includegraphics[width=2.8cm]{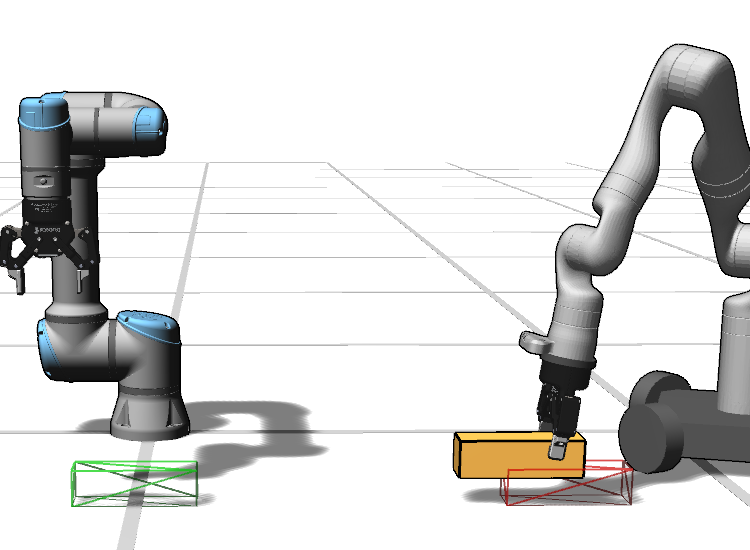}};
            \draw (0, 1.0) node[align=center] {7 s};
        \end{tikzpicture}
    \end{subfigure}
    \caption{When the manipulators are separated by a wall the robots must cooperate in order to move the block from the start to the goal pose (outlined in green and red). When the wall is removed the block can be moved from the start to the goal pose by a single robot.}
    \label{fig:wall_anim}
    \vspace{-1em}
\end{figure}
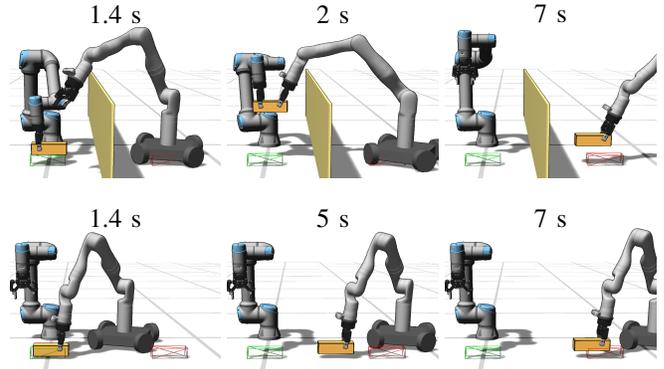

Fig. \ref{fig:wall_anim} shows key points of the emerging trajectories.
The wall, when present, is placed such that it prevents the Kinova platform from directly reaching for the package and transporting it to the goal position.
The robots may thus cooperate, and in the resulting trajectory the block is first lifted by the UR5 and handed over to the Kinova platform, which brings the block to the goal position.
When the wall is absent (all other parameters being equal) the task is completed by only the Kinova platform.

\subsection{Multiple moving objects}
This experiment consists of three blocks that need to be moved, and four stationary manipulators arranged in a rectangular pattern between the start and goal positions of the blocks.
In this experiment we use a trajectory consisting of 13 segments.

\begin{figure}
    \centering
    \includegraphics[width=\columnwidth]{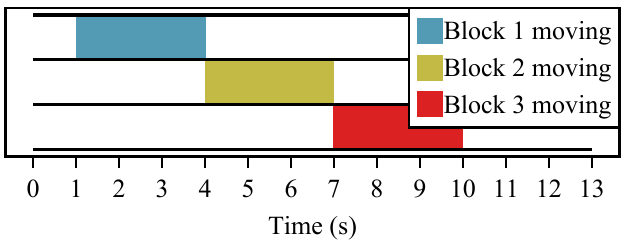}
    \caption{Tasks involving multiple objects benefit from initialization. We construct the initialization by linearly interpolating the poses of the movable objects within non-overlapping time windows.}
    \label{fig:exp2_init}
    \vspace{-1em}
\end{figure}

\begin{figure}
    \vspace{-1em}
    \centering
    \includegraphics[width=\columnwidth]{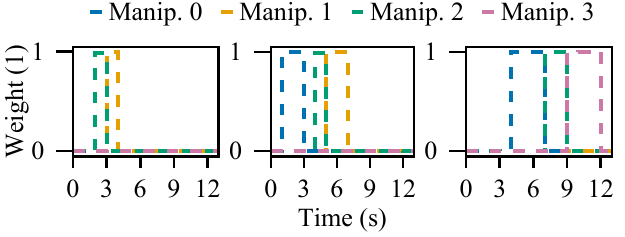}
    \caption{Association weights between the manipulators and the blocks as a function of time. At time 3-4 s the middle block is placed on the ground before being picked up again.}
    \label{fig:multiple_packages_control}
\end{figure}

The initialization is created by linearly interpolating the trajectories of the blocks and distributing them over time.
An illustration of the heuristic schedule is shown in Fig. \ref{fig:exp2_init}.
We then execute ten iterations of the Gauss-Newton solver while keeping the trajectories of the blocks fixed.
The resulting manipulator trajectories are then used for initializing the actual optimization where the trajectories of the blocks are included.
\begin{figure}
    \begin{subfigure}[b]{0.5\columnwidth}
        \centering
        \begin{tikzpicture}
            \draw (0, 0) node {\includegraphics[width=4.1cm]{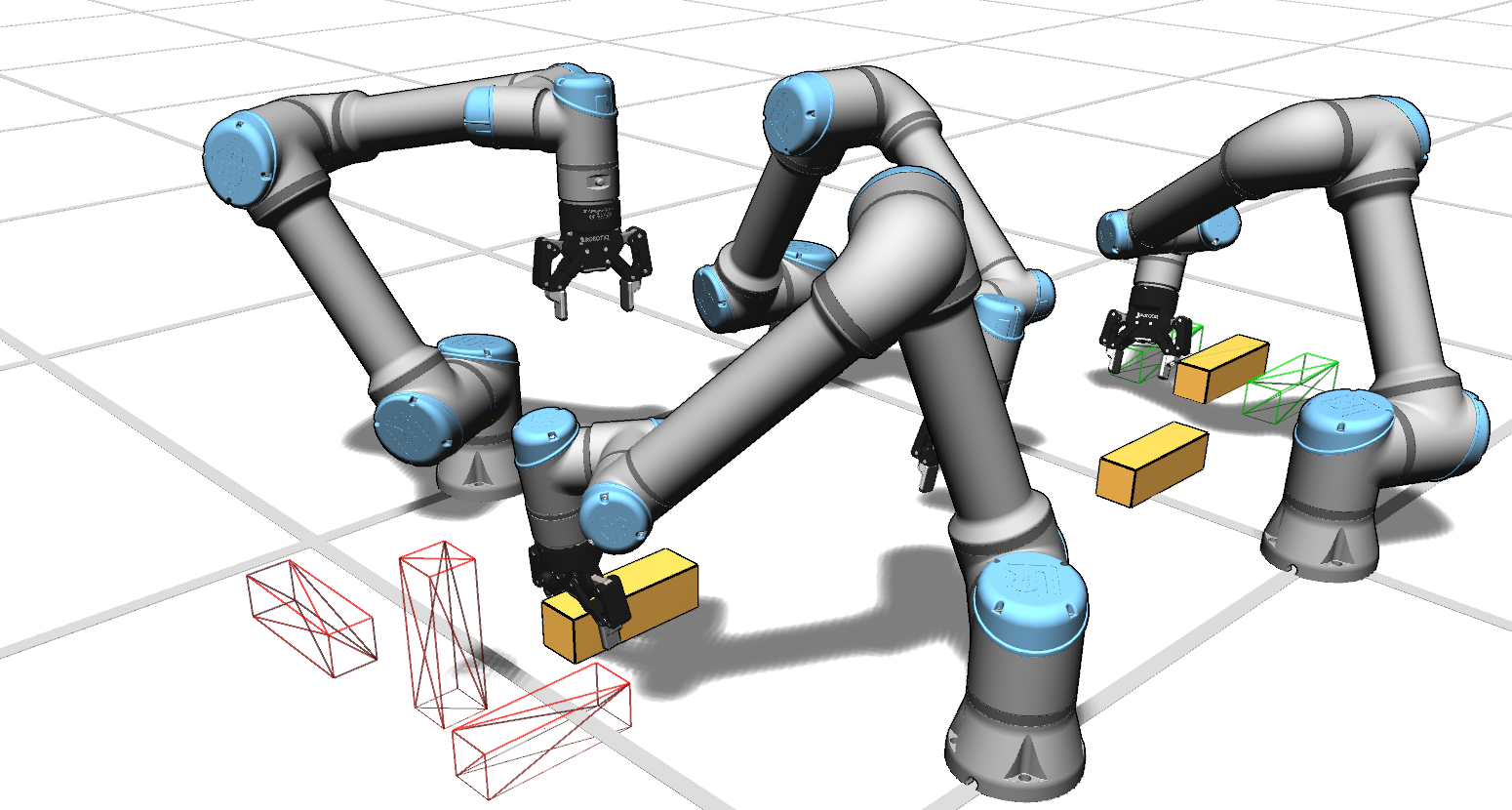}};
            \draw (0, 1.35) node[align=center] {3.5 s};
        \end{tikzpicture}
    \end{subfigure}%
    \hfill
    \begin{subfigure}[b]{0.5\columnwidth}
        \centering
        \begin{tikzpicture} 
            \draw (0, 0) node {\includegraphics[width=4.1cm]{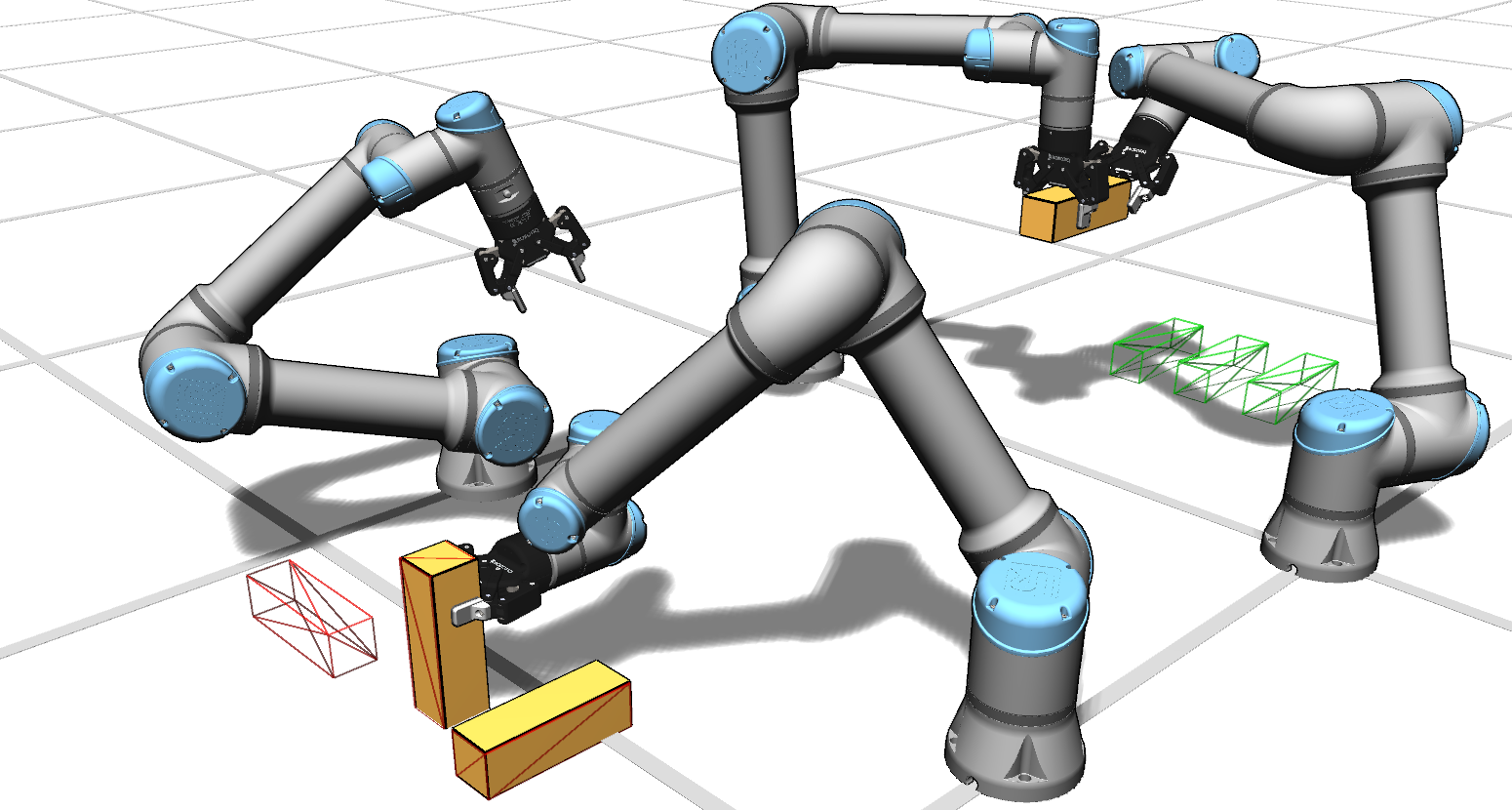}};
            \draw (0, 1.35) node[align=center] {7 s};
        \end{tikzpicture}
    \end{subfigure}
    \caption{The second experiment requires multiple handovers for successful completion.}
    \label{fig:multiple_packages}
    \vspace{-1.0em}
\end{figure}
The solution is captured in Fig. \ref{fig:multiple_packages}.
Noteworthy is that the solution contains a segment where one of the blocks is placed to rest on the ground before being picked up again and transported to the goal position.
The association weights for the blocks are visualized in Fig. \ref{fig:multiple_packages_control}, showing one block being at rest at time 3-4 s.

\subsection{Including interactive objects}\label{sec:result_drawer}
In the previous experiments the starting pose of the blocks are directly reachable by one or more manipulators.
However, oftentimes the object of interest can be reached only after manipulating the environment, such as by opening a drawer.
Interactive objects can be directly included into \eqref{eq:trajectory_problem} as agents subject to appropriate constraints.

\begin{figure}
    \begin{subfigure}[b]{0.5\columnwidth}
        \centering
        \begin{tikzpicture}
            \draw (0, 0) node {\includegraphics[trim=200 160 200 19,clip,height=3.4cm]{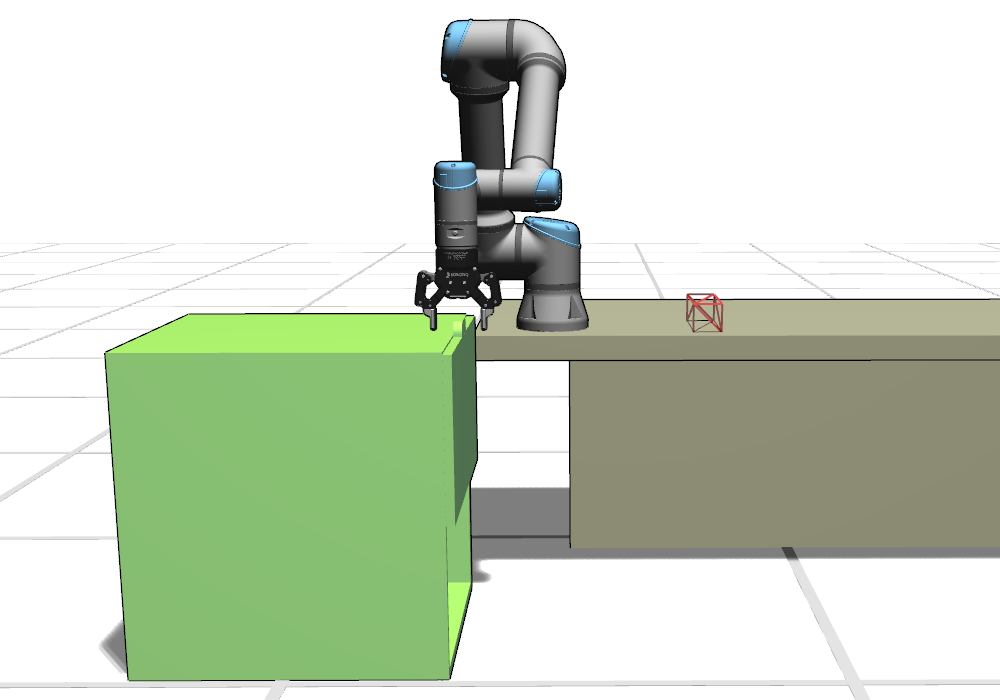}};
            \draw (-1.0, 1.5) node[align=center] {2 s};
        \end{tikzpicture}
    \end{subfigure}%
    \hfill
    \begin{subfigure}[b]{0.5\columnwidth}
        \centering
        \begin{tikzpicture} 
            \draw (0, 0) node {\includegraphics[trim=200 160 200 19,clip,height=3.4cm]{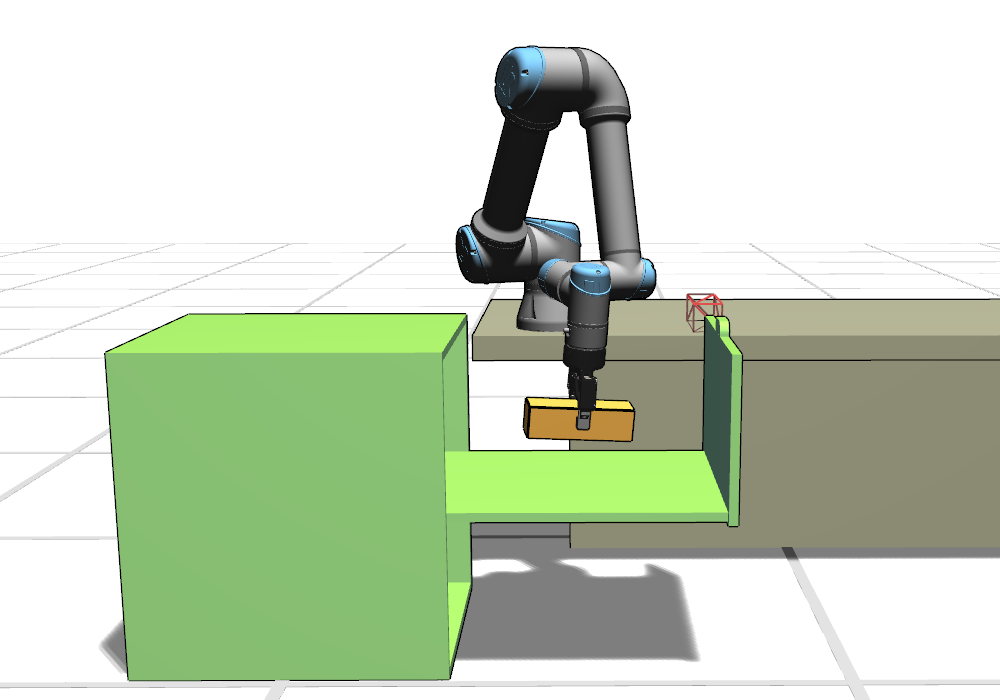}};
            \draw (-1.0, 1.5) node[align=center] {6.5 s};
        \end{tikzpicture}
    \end{subfigure}
    \caption{The formulation also supports interactive objects. Here the drawer needs to be opened before the block can be retrieved.}
    \label{fig:drawer_trajectory}
    \vspace{-1.0em}
\end{figure}

We include a drawer by treating it simultaneously as an object that can be moved and as a manipulator equipped with an end effector that can hold objects.
The drawer is thus subject to a velocity constraint which allows it to move only when actuated by another manipulator, and a corresponding pose constraint attached to the handle of the drawer (see \eqref{eq:velocity_constraint} and \eqref{eq:position_constraint}).
The start pose of the block is inside of the closed drawer while the goal pose is on the table next to the robot as shown in Fig. \ref{fig:drawer_trajectory}.
We initialize the trajectory of the block by linearly interpolating between the start and the goal poses.
The trajectory of the drawer is initialized to the resting pose, i.e. closed, while the trajectory of the UR5 is initialized to the rest state.
The association weight between the drawer and the block is initialized to 1.
The resulting trajectory features the UR5 opening the drawer before reaching for the block and placing it on the table.

\subsection{Extension for multiple grasping orientations}
The the formulation in \ref{sec:detailed_method} provides some flexibility in choosing the grasp pose through the longitudinal offset $\delta_{ij}$ and the angle $\theta_{ij}$, but the grasping orientation is still fixed around the two remaining axes of the object.
However, the blocks used in the previous examples allow four distinct grasping orientations around the longitudinal axis.
In this section we show how the formulation in \ref{sec:detailed_method} can be extended in order to enable all of these orientations.

We introduce the functions $\gamma_{ijk}:[0,T]\to [0,1]$ that denote the association of a particular grasping orientation $k$ between an object $j$ and a manipulator $i$. The position constraints \eqref{eq:position_constraint} can now be replaced with
\begin{equation*}
    \gamma_{ijk}(t)w_{ij}(t)\left(\mankin_i(\m_i) - \hat{\objkin}_{k}(\p_j,\y_{ij})\right) = \0,
\end{equation*}
where $\hat{\objkin}_{k}$ is the target grasp pose after applying the offset corresponding to index $k$.
By also setting ${\sum_k \gamma_{ijk}(t)=1\; \forall \; i\in\mathcal{M}}$
we can ensure that at least one alternative will be active.

\begin{figure}
    \begin{subfigure}[b]{0.5\columnwidth}
        \centering
        \begin{tikzpicture}
            \draw (-0.2, 0) node {\includegraphics[trim={0 0 0 70}, clip, width=4.1cm]{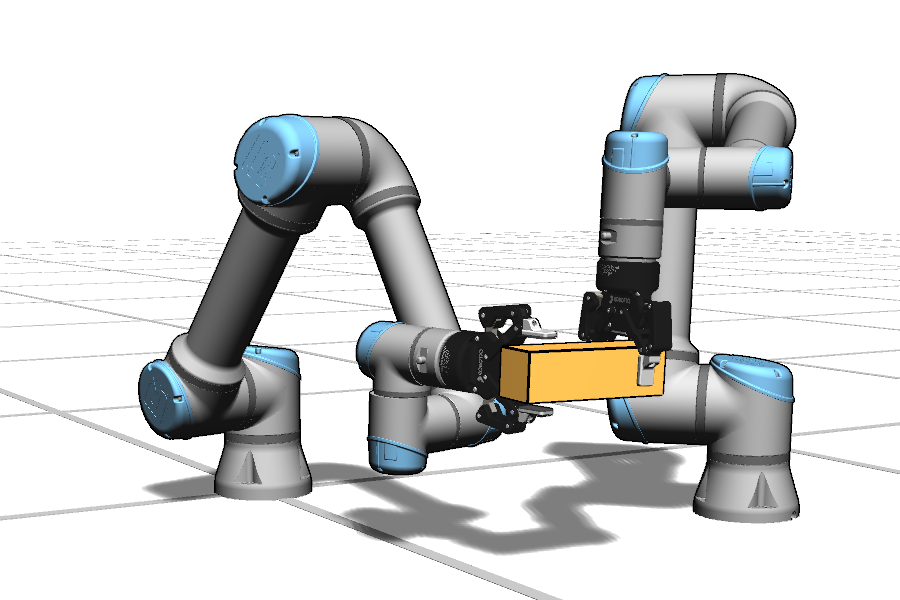}};
            \draw (0, 0.8) node[align=center] {2 s};
        \end{tikzpicture}
    \end{subfigure}%
    \hfill
    \begin{subfigure}[b]{0.5\columnwidth}
        \centering
        \begin{tikzpicture} 
            \draw (0.0, 0) node {\includegraphics[trim={0 0 0 70}, clip, width=4.1cm]{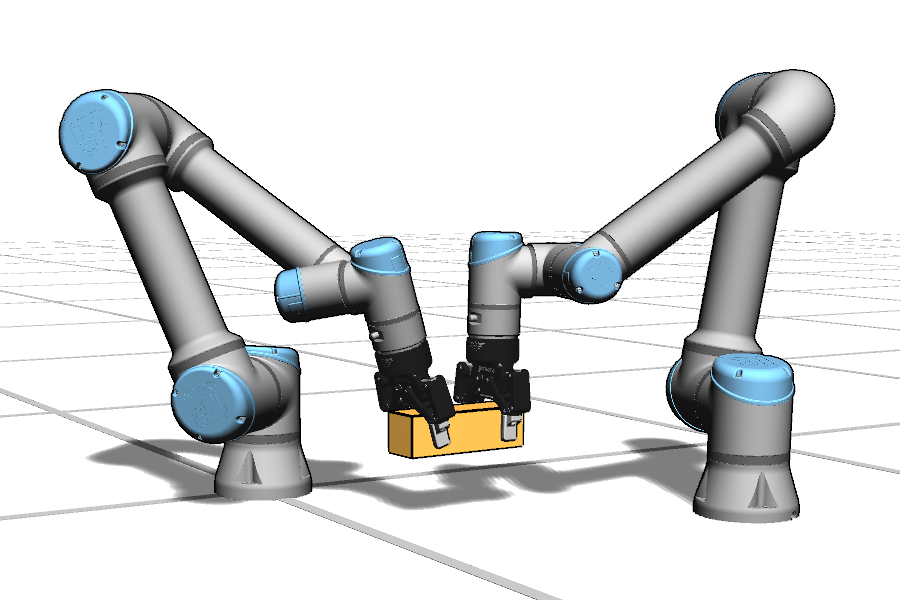}};
            \draw (0, 0.8) node[align=center] {7.0 s};
        \end{tikzpicture}
    \end{subfigure}
    \begin{subfigure}[b]{\columnwidth}
        \includegraphics[width=\columnwidth]{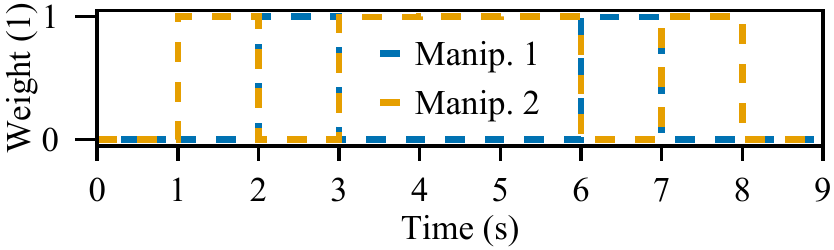}
    \end{subfigure}
    \caption{Here two robots are reorienting a block such that the face that is initially facing upwards will be facing towards the ground.}
    \label{fig:exp_reorientation}
\end{figure}

\begin{figure}
    \centering
    \includegraphics[width=\columnwidth]{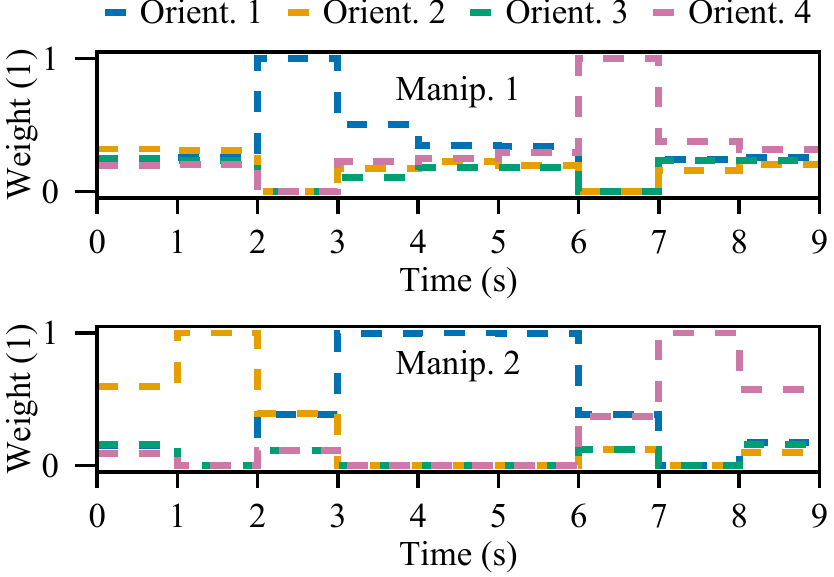}
    \caption{The orientation weights $\gamma_{ijk}(t)$ as a function of time. Manipulator 1 is grasping the block from two different directions while manipulator 2 is using three different ones. The constraint becomes active when both $\gamma_{ijk}(t)$ and $w_{ij}(t)$ are greater than zero. }
    \label{fig:exp_reorientation_or_weights}
    \vspace{-1.0em}
\end{figure}

The updated formulation can be used to solve e.g. reorientation problems.
In Fig. \ref{fig:exp_reorientation}, two manipulators are tasked to reorient a block such that the face that is initially facing upwards will be facing towards the ground at the end.
The values of $\gamma_{ijk}$ are shown in Fig. \ref{fig:exp_reorientation_or_weights}.
The reorientation of the block could in this case be completed with only one handover, however, the obtained solution is a local minimum featuring four handovers.
In this experiment the trajectory of the block has been initialized to a linearly interpolated trajectory between the start and the goal pose while the manipulator trajectories are initialized to the rest pose.

\subsection{Weight derivative limit}
As discussed in \ref{sec:constraint_manifold}, the formulation enables multiple robots to seamlessly transport a single package. We can induce this behavior by introducing an additional constraint that limits the rate of change of the association weights $w_{ij}$. This constraint is of the form
\begin{align*}
    C_{\Delta^+_w}:= \dot{w}_{ij} \leq \Delta_w^+\\
    C_{\Delta^-_w}:= \dot{w}_{ij} \geq \Delta_w^-
\end{align*}
where $\Delta_w^+$ and $\Delta_w^-$ denote the upper and lower limit of the derivatives.

\begin{figure}
    \begin{subfigure}[b]{0.5\columnwidth}
        \centering
        \begin{tikzpicture}
            \draw (0, 0) node {\includegraphics[trim={0 0 0 0}, clip, width=4.1cm]{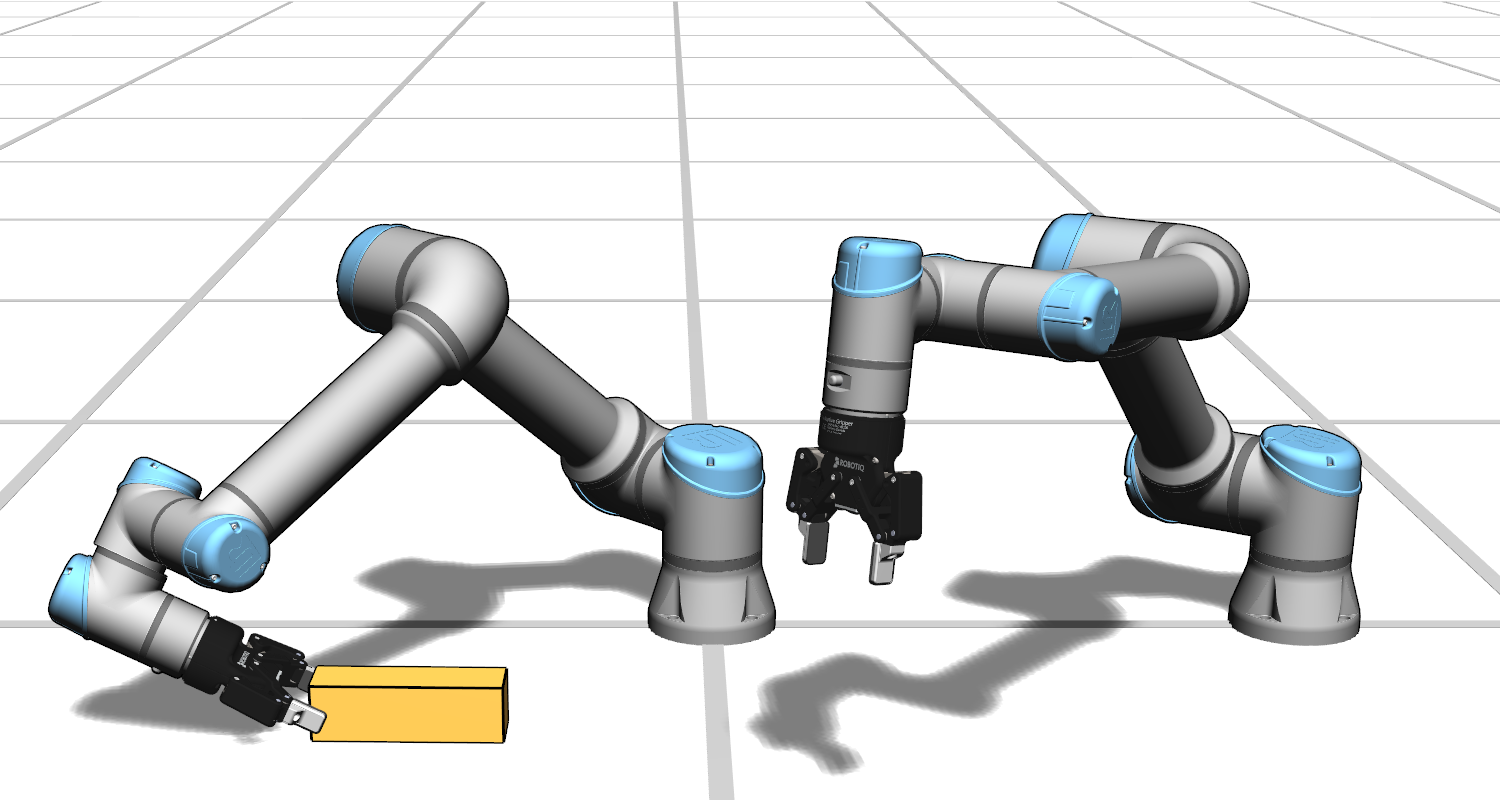}};
            \draw (0.2, 1.0) node[align=center] {2.5 s};
        \end{tikzpicture}
    \end{subfigure}%
    \hfill
    \begin{subfigure}[b]{0.5\columnwidth}
        \centering
        \begin{tikzpicture} 
            \draw (0.0, 0) node {\includegraphics[trim={0 0 0 0}, clip, width=4.1cm]{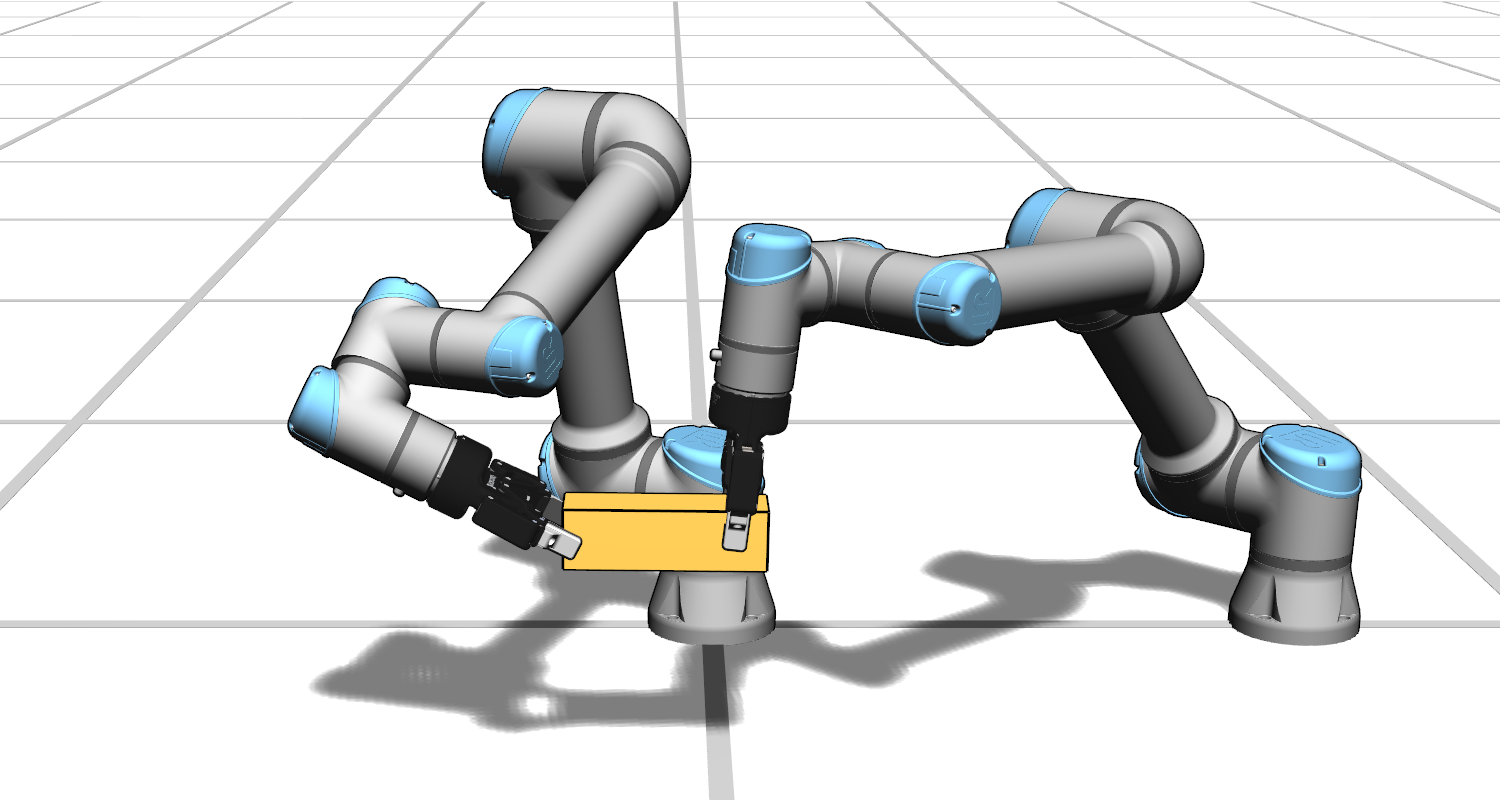}};
            \draw (0.2, 1.0) node[align=center] {3.5 s};
        \end{tikzpicture}
    \end{subfigure}
    \begin{subfigure}[b]{\columnwidth}
        \includegraphics[width=\columnwidth]{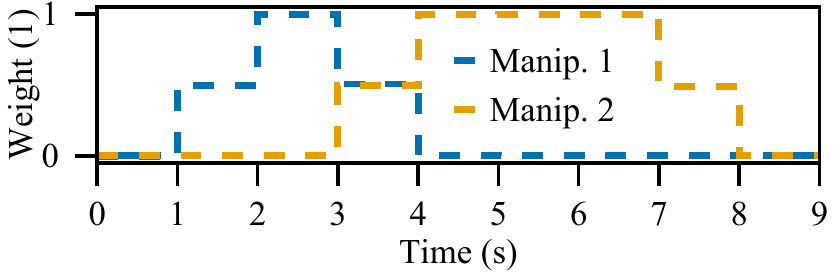}
    \end{subfigure}
    \caption{Two UR5s performing a handover where the switching time is constrained. During time segments 1-2 s and 7-8 s, the sum of the association weights is less than 1, and as discussed in \ref{sec:constraint_manifold}, the velocity constraint \eqref{eq:velocity_constraint} then ensures that the object stays in place.}
    \label{fig:exp_w_limit}
    \vspace{-1.0em}
\end{figure}

This constraint causes the pick-up and release phases to be extended. The weights now need $\frac{1}{\Delta_w^+}$ time units to switch from 0 to 1, which can be useful e.g. in order to provide enough time for the grippers to open and close during a pick-up or a handover. As discussed in \ref{sec:constraint_manifold}, the velocity can be non-zero only when the weights of one object sum up to one, and therefore the velocity of both the end effector and the object will be zero when the object is picked up and dropped on the ground. During a handover the object may still move as long as the weights sum up to 1. An experiment with two UR5s where the upper and lower derivative limits are set to $\frac{1}{2}$ and $-\frac{1}{2}$ respectively is shown in Fig. \ref{fig:exp_w_limit}.

\subsection{Optimization runtime}
\begin{figure}
     \begin{subfigure}[b]{0.29\columnwidth}
         \centering
         \includegraphics[trim={80 0 10 10}, clip, width=\textwidth]{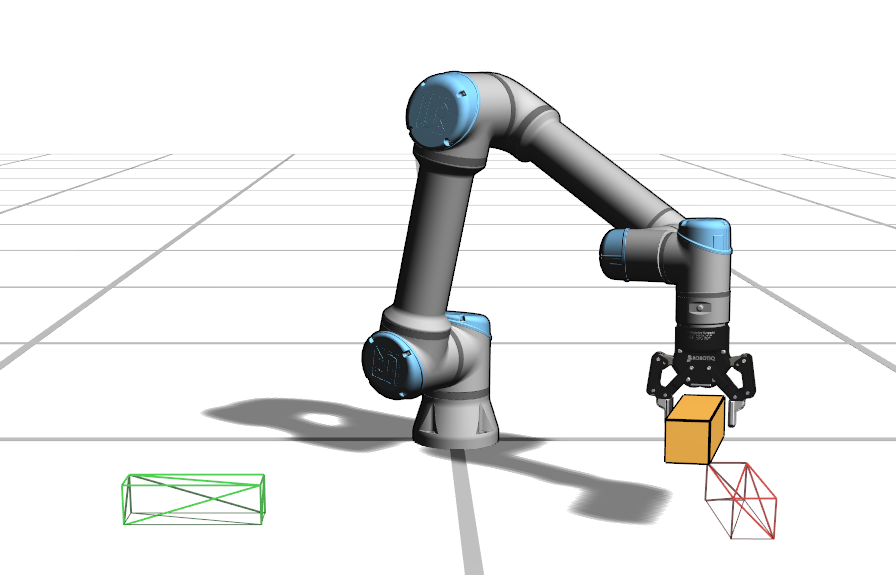}
     \end{subfigure}
     \begin{subfigure}[b]{0.69\columnwidth}
         \centering
         \includegraphics[trim={50 0 0 10}, clip, width=\textwidth]{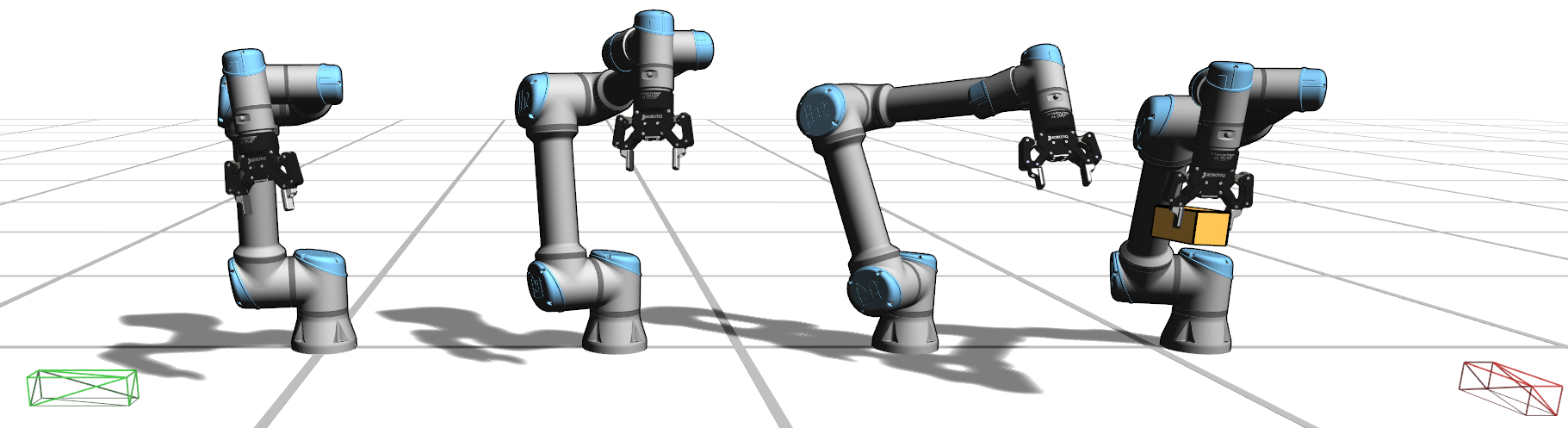}
     \end{subfigure}
     \begin{subfigure}[b]{0.39\columnwidth}
         \centering
         \includegraphics[trim={180 0 160 0}, clip, width=\textwidth]{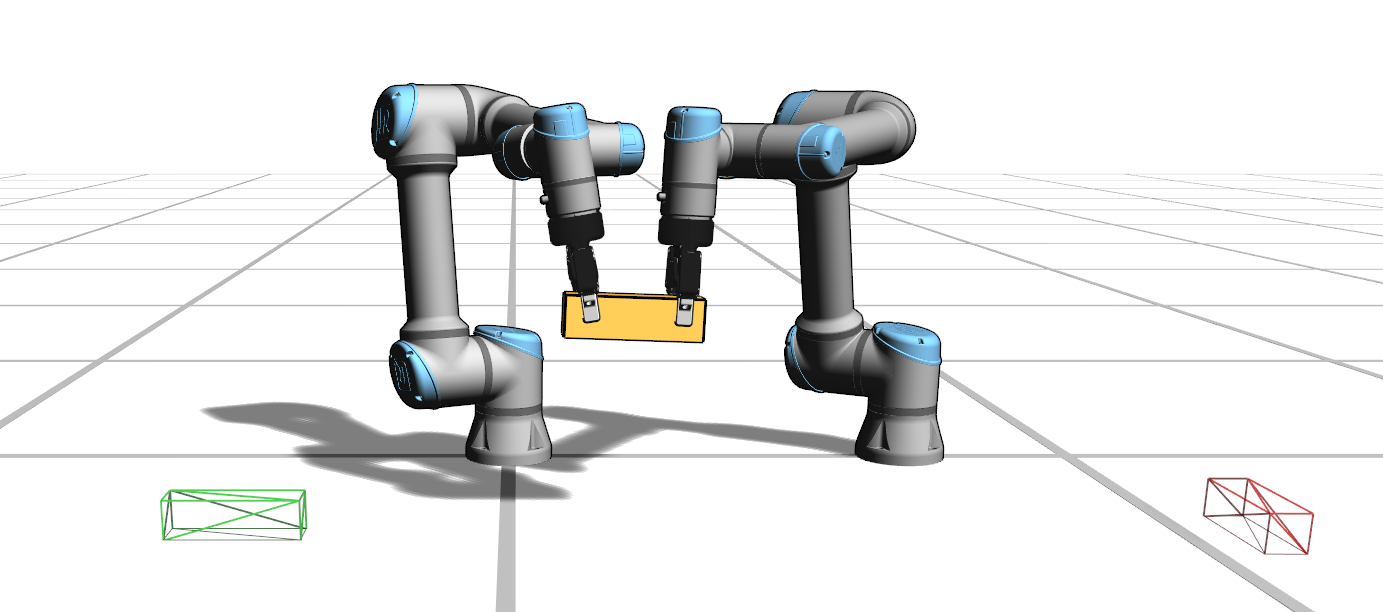}
     \end{subfigure}
     \begin{subfigure}[b]{0.59\columnwidth}
         \centering
         \includegraphics[trim={100 10 130 0}, clip, width=\textwidth]{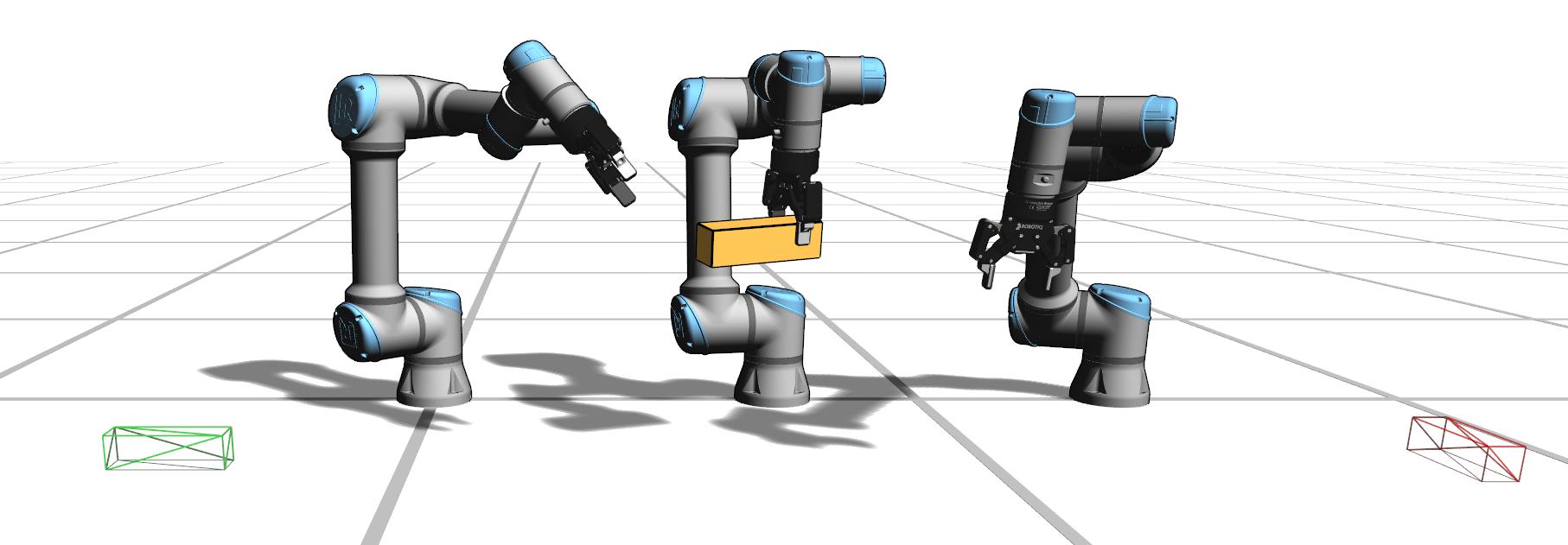}
     \end{subfigure}
    \begin{subfigure}[b]{\columnwidth}
        \vspace{0.5em}
        \centering
        \includegraphics[width=\columnwidth]{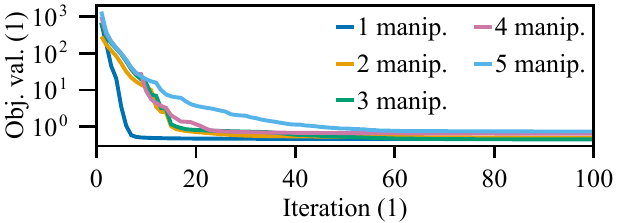}
    \end{subfigure}
    \caption{The objective value as a function of the iteration number with different numbers of manipulators.} %
    \label{fig:convergence}
    \vspace{-1.5em}
\end{figure}

Finally we study how increasing the number of manipulators impacts the optimization. In this experiment we assemble UR5s on a line with the task of moving a block from the start of the line to the end.
Adding manipulators increases the number of iterations needed for convergence, as shown in Fig. \ref{fig:convergence}. The five manipulator setup is shown in Fig. \ref{fig:teaser}. The runtime and the number of variables for each experiment is shown in \ref{tab:runtimes}. The table must be interpreted carefully as the exact task description has a significant impact on the optimization landscape and thus also the number of iterations needed for convergence. The measurements reported in \ref{tab:runtimes} also include rendering and are intended as rough estimates only.
\begin{table}[t]
\centering
\caption{Number of variables and average runtime for different number of manipulators. The reported runtime is the average of four measurements.}
\label{tab:runtimes}
\begin{tabular}{c|ccccc}
\hline
Manipulators        & 1   & 2    & 3    & 4    & 5    \\ \hline
Wall time (s)       & 9.2 & 20.3 & 28.5 & 36.5 & 46.3 \\
Number of variables & 305 & 436  & 567  & 698  & 829  \\ \hline
\end{tabular}
\vspace{-1.0em}
\end{table}
\section{Discussion and Conclusion}
\label{sec:discussion}

The results in \ref{sec:results} show that decisions regarding high level actions such as pick, place and open can be obtained implicitly by solving a nonlinear optimization problem.
The actions emerge automatically as part of the solution without the need to manually specify which manipulator should be working, or even when.

Trajectory optimization involving collision avoidance is in general a non-convex problem.
Gradient based methods use only local information and may therefore struggle in finding new trajectories that lie far away from the initial guess.
This can be demonstrated by constructing a variant of the drawer experiment from \ref{sec:result_drawer} where the goal position of the block is placed on top of the drawer which did not converge to any reasonable solution in our experiments.
This problem has also been identified in \cite{bib:takano_continuous_2021}.
Finding the global optimum of the problem requires more sophisticated optimization algorithms that are capable of exploring the optimization space efficiently.
Existing TAMP algorithms may in some cases be able to work around this problem by including the actions themselves into the problem formulation, turning the actions into conditions for solving the optimization problem.

Additionally it is not entirely clear how the formulation could be extended to support objects with inherently discrete states, e.g. light switches, while still being continuous.
Our formulation also does not have any notion of temporal precedence, i.e. objects may arrive at the goal position in an arbitrary order.
This can, however, be mitigated to some extent by careful initialization.

We believe that the formulation presented here can be useful, especially as part of a larger TAMP algorithm.
Even when combined with existing TAMP algorithms, by finding some actions implicitly it would be possible to reduce the number of actions that must be considered during the sequencing.
As the optimization method used in this work might have difficulties in highly non-convex problems we would additionally like to investigate the use of sampling based methods for handling the end effector constraints.

\bibliographystyle{IEEEtran}
\bibliography{references.bib,morereferences.bib}

\end{document}